# Information Transfer in Swarms with Leaders

YU SUN, LOUIS F. ROSSI and CHIEN-CHUNG SHEN, University of Delaware
JENNIFER MILLER, Trinity College
X. ROSALIND WANG, JOSEPH T. LIZIER and MIKHAIL PROKOPENKO, CSIRO Computational Informatics
UPUL SENANAYAKE, The University of Sydney

## 1. INTRODUCTION

Swarm dynamics is the study of collections of agents that interact with one another without central control. In natural systems, insects, birds, fish and other large mammals function in larger units to increase the overall fitness of the individuals. Their behavior is coordinated through local interactions to enhance mate selection, predator detection, migratory route identification and so forth [Andersson and Wallander 2003; Buhl et al. 2006; Nagy et al. 2010; Partridge 1982; Sumpter et al. 2008]. In artificial systems, swarms of autonomous agents can augment human activities such as search and rescue, and environmental monitoring by covering large areas with multiple nodes [Alami et al. 2007; Caruso et al. 2008; Ogren et al. 2004; Paley et al. 2007; Sibley et al. 2002]. In this paper, we explore the interplay between swarm dynamics, covert leadership and theoretical information transfer. A leader is a member of the swarm that acts upon information in addition to what is provided by local interactions. Depending upon the leadership model, leaders can use their external information either all the time or in response to local conditions [Couzin et al. 2005; Sun et al. 2013]. A covert leader is a leader that is treated no differently than others in the swarm, so leaders and followers participate equally in whatever interaction model is used [Rossi et al. 2007]. In this study, we use theoretical information transfer as a means of analyzing swarm interactions to explore whether or not it is possible to distinguish between followers and leaders based on interactions within the swarm. We find that covert leaders can be distinguished from followers in a swarm because they receive less transfer entropy than followers.

## 2. MODELING THREE-ZONE SWARMING WITH COVERT LEADERS

In swarms of autonomous individuals, large-scale behavior results from decisions made based on local information. Local interactions are often defined by three zones surrounding each individual [Aoki 1982; Couzin et al. 2002; Huth and Wissel 1992; Lukeman et al. 2010; Vicsek et al. 1995]. The zone closest to the individual is the zone of repulsion, and the individual will respond to neighbors in this zone by moving away from them. The zone of orientation is further away, and an individual will align itself with the direction of travel for neighbors in this region. Finally, an individual will move toward neighbors in its zone of attraction. The responses to all neighbors are combined, and the individual turns toward this desired direction.

We use continuous zones of interaction to avoid having an individual's desired direction make a sudden large change. The responses to neighbors are weighted by Gaussian kernels depending on the distance from the individual to its neighbors. Although the kernels take every individual in the swarm into account, they decay so quickly that the influence of a far-off individual is negligible compared to closer neighbors. A cross-section view of the kernels is shown in Fig. 1.

The individuals in our swarms are governed by a system of differential equations introduced in [Miller et al. 2012]. For an individual, we define the position vector to be $\vec{s}$. To compute the local interactions for individual $i$, we use $\vec{s}_{ij} = \vec{s}_j - \vec{s}_i$ for the distance between neighbors $i$ and $j$. The desired velocity vector for individual $i$ is given by $\vec{v}_{d,i} = \vec{v}_{r,i} + \vec{v}_{o,i} + c_a \vec{v}_{a,i}$ where :

$$\vec{v}_{r,i} = \sum_{j=1}^{N} -\frac{1}{8\pi\sigma_1^4} \vec{s}_{ij} \exp(-|\vec{s}_{ij}|^2/4\sigma_1^2), \quad \vec{v}_{o,i} = \frac{\sum_{j=1}^{N} \frac{1}{4\pi\sigma_2^2} \exp(-|\vec{s}_{ij}|^2/4\sigma_2^2)\vec{v}_j}{\sum_{j=1}^{N} \frac{1}{4\pi\sigma_2^2} \exp(-|\vec{s}_{ij}|^2/4\sigma_2^2)},$$

$$\vec{v}_{a,i} = \sum_{j=1}^{N} \frac{1}{64\pi\sigma_3^6} \vec{s}_{ij} |\vec{s}_{ij}|^2 \exp(-|\vec{s}_{ij}|^2/4\sigma_3^2), \quad (1)$$

where $c_a$ controls the relative importance of attraction, and $\sigma_1$, $\sigma_2$, and $\sigma_3$ control the zone widths. Once $\vec{v}_d$ has been computed, the velocity for each individual is updated.

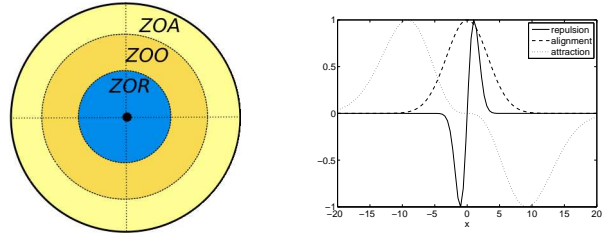

Fig. 1. Location of zones of repulsion (ZOR), orientation (ZOO) and attraction (ZOA) along with cross-section views of normalized interaction kernels.

We use the methodology in [Sun et al. 2013] to model covert leaders in the swarm. In this case, the additional information is a preferred direction for the swarm. We use a modified form of the linear leadership model [Couzin et al. 2005] so that leaders will respond more strongly to their additional information where the swarm is sparse (low density), and interact more like followers where the swarm is denser. In our model, the follower's desired velocity $\vec{v}_{lf}$ is the same as for the leaderless model (1). The desired velocity for a covert leader $\vec{v}_{ld}$ is represented by the following equation:

$$\vec{v}_{ld} = \left(1 - e^{-\frac{G_{\sigma_2} * (\rho_f + \rho_l)}{\sigma}}\right) \vec{v}_{lf} + e^{-\frac{G_{\sigma_2} * (\rho_f + \rho_l)}{\sigma}} \vec{g}, \quad (2)$$

where the $*$ denotes a convolution, $G$ is Gaussian smoothing kernel and $\rho_f$ and $\rho_l$ is the follower and leader density. The leader's influence decays over the scale $\sigma$. Our analysis in [Sun et al. 2013] shows that the stability properties are the same as for the leaderless model so information can be inserted into the system without changing the stability of the coherent structure.

## 3. TRANSFER ENTROPY FOR SWARMS

We utilized a framework for local information transfer developed by Lizier et al. [Lizier et al. 2008]. The framework precisely quantifies information transfer at each spatiotemporal point in a complex system. Previous works [Wang et al. 2012; Miller et al. 2014] showed long range communications between individuals within a swarm can be captured by conditional transfer entropy (CTE). The local CTE from a source agent $Y$ to a destination agent $X$ conditioned on another contributor $W$ is the local mutual information between the previous state of the source $y_n$ and the next state of the destination $x_{n+1}$, *conditioned* on the past of the destination $x_n^{(k)}$ and the previous state of the contributor $w_n$:

$$t_{Y \to X|W}(n+1, k) = \lim_{k \to \infty} \log_2 \frac{p(x_{n+1}|x_n^{(k)}, w_n, y_n)}{p(x_{n+1}|x_n^{(k)}, w_n)}, \quad (3)$$

where $t_{Y \to X}(n+1, k)$ represents finite-$k$ approximation. The variables in Equation 3 are composed of the relative positions and change in velocity of the swarm individuals:

$$y_n = \{\vec{s}_p^n - \vec{s}_{p'}^n, \vec{v}_p^n - \vec{v}_{p'}^n\}, \quad w_n = |v|^n, \quad x_n = \vec{v}_p^n - \vec{v}_p^{n-1}, \quad x_{n+1} = \vec{v}_p^{n+1} - \vec{v}_p^n$$

where $p$ is the destination swarm individual and $p'$ is the source individual.

To calculate the CTE in swarms, previous works [Wang et al. 2012; Miller et al. 2014] treated the swarm as homogeneous and estimated the probability distribution function (PDF) in Eq. 3 by accumulating the observations across all agents. In this paper however, the leaders and followers are known to react differently given their neighboring conditions, and thus the PDF needs to capture the reactions of the individuals within the swarm accordingly. Therefore, we separate the observations by the role of the destination individual, and estimate separate PDFs for those destination individuals that are followers and those that are leaders. We characterize overall transfer as the average over all causally connected pairs $Y \to X^F$, where $F$ denotes followers, at each time step: $T(n+1, k)^F = \langle t_{Y \to X^F}(n+1, k) \rangle_{Y \to X^F}$, and similarly for the leaders.

## 4. RESULTS: INFORMATION THEORY AND IDENTIFYING COVERT LEADERS

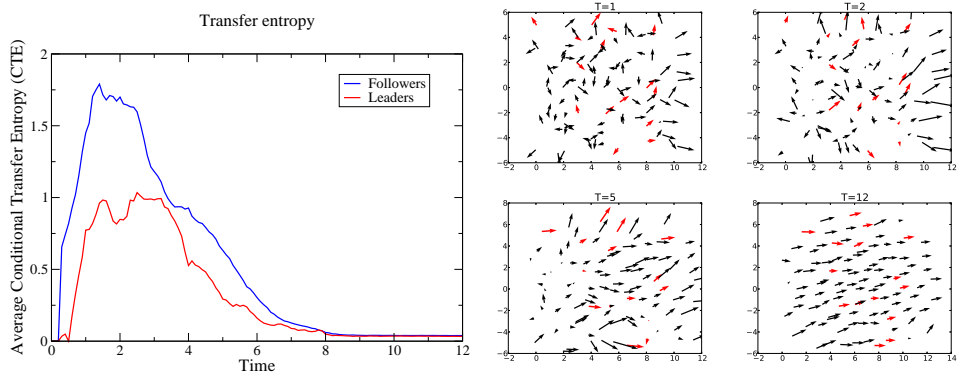

Fig. 2. The transfer entropy received by followers and leaders over time along with the swarm configuration at key times during self-organization. Leaders are displayed in red, but swarm interactions with leaders and followers are the same.

Though it can be difficult, if not impossible, to identify leaders through visual inspection, information theory provides an appropriate lens through which this segregation may be possible. (In certain models under certain circumstances, leaders may aggregate at the front, but this is not generally the case.) Using the information theoretic approach, we discard all a priori knowledge of individual interactions and instead use the CTE in the swarm to see if covert leaders transmit or receive more information than followers. While it is intuitive that a leader ought to share more information with the swarm, we find that the reciprocal relationship is the best way to distinguish covert leaders. *A covert leader in a swarm is notable because it receives less information than followers on average.* In Fig. 2, we can see an experiment where a regular array of randomly oriented individuals organizes itself into a coherent translating disk. Without leadership, the equilibrium direction of motion is random, depending upon the initial disordered configuration. In this experiment, 15% of the individuals are leaders that want to move to the right ($\vec{g} = [1, 0]^T$). While it is not possible to identify the leaders through visual inspection, the average CTE received by the covert leaders is markedly lower than the CTE received by the followers as the swarm organizes itself into a coherent disk.